\documentclass{article}
\usepackage{spconf,amsmath,graphicx,hyperref}
\usepackage{cite}
\usepackage{amsmath,amssymb,amsfonts}
\usepackage{algorithmic}
\usepackage{graphicx}
\usepackage{textcomp}
\usepackage{xcolor}
\usepackage{booktabs}
\usepackage{makecell}
\usepackage{pifont}
\usepackage{newunicodechar}
\usepackage{appendix}
\newunicodechar{✓}{\ding{51}}
\newunicodechar{✗}{\ding{55}}
\def\BibTeX{{\rm B\kern-.05em{\sc i\kern-.025em b}\kern-.08em
    T\kern-.1667em\lower.7ex\hbox{E}\kern-.125emX}}

\title{Evaluating and Improving the Robustness of Speech Command Recognition Models to Noise and Distribution Shifts}
\twoauthors
  {Anaïs BARANGER}
	{Thales SIX\\
	MultiMedia Processing\\
	Gennevilliers, France}
  {Lucas MAISON\sthanks{Work performed while at Thales SIX.}}
	{Avignon University\\
	Laboratoire Informatique d'Avignon\\
	Avignon, France}
\begin{document}
%
\maketitle
\begin{abstract}
Although prior work in computer vision has shown strong correlations between in-distribution (ID) and out-of-distribution (OOD) accuracies, such relationships remain underexplored in audio-based models. In this study, we investigate how training conditions and input features affect the robustness and generalization abilities of spoken keyword classifiers under OOD conditions.
We benchmark several neural architectures across a variety of evaluation sets. To quantify the impact of noise on generalization, we make use of two metrics: Fairness (F), which measures overall accuracy gains compared to a baseline model, and Robustness (R), which assesses the convergence between ID and OOD performance. Our results suggest that noise-aware training improves robustness in some configurations. These findings shed new light on the benefits and limitations of noise-based augmentation for generalization in speech models.
\end{abstract}
\section{Introduction}
\label{sec:intro}

In recent years, deep learning models have achieved remarkable performance in Automatic Speech Recognition (ASR) and Spoken Command Recognition (SCR) tasks. However, those models are often evaluated on test data that follows the same distribution as the training data. When facing out-of-distribution (OOD) conditions, such as new speakers or noisy environments, the performance of these models may be degraded significantly. In computer vision, several studies have observed a correlation between in-distribution (ID) and OOD accuracy, a phenomenon often called "\emph{accuracy-on-the-line}"~\cite{pmlr-v139-miller21b, baek2022agreement}. This property suggests that improving ID accuracy also improves OOD accuracy. However, 
there is limited evidence supporting or refuting this correlation in audio models. Some studies address OOD accuracy in spoken language understanding~\cite{Porjazovski_2024}, introducing OOD generalization aspects such as novel vocabulary, new semantic combinations, and acoustic variations. Other studies focus on the robustness of ASR models across various demographic subgroups~\cite{DBLP:journals/corr/abs-2403-07937, rakib2023ood}. However, these works rarely focus on the specific effect of noisy training for speech recognition in diverse environments, such as different speakers or various types of noise, and even less on the impact of training conditions and model architecture on robustness. Only for computer vision tasks has it been shown that the introduction of noise or nuisance factors can break the correlation mentioned above, a phenomenon the authors coined "\emph{accuracy-on-the-wrong-line}"~\cite{sanyal2024accuracy}. Whether a similar breakdown occurs in speech recognition models is an open question.

In this work, we address two main research objectives. First, we analyze the effect of noise on the ID–OOD accuracy correlation in SCR models. Previous works~\cite{pmlr-v139-miller21b, baek2022agreement} demonstrated a strong ID–OOD correlation, but focused mainly on clean datasets and vision tasks. We aim to verify whether this relationship holds in realistic and noisy audio settings. In particular, we test whether the hypothesis proposed in~\cite{sanyal2024accuracy} (that the presence of noise can shatter the \emph{accuracy-on-the-line} phenomenon) is true for the command recognition task. Second, we systematically assess and quantify the impact of noisy training data or input features on the OOD robustness of speech models. 
To conduct these experiments, we chose to use TDNN~\cite{snyder2019speaker}, CNN, and DNN~(with and without recurrence) architectures because these models represent a diverse set of approaches widely used in speech classification tasks. TDNNs are known for their temporal modeling capabilities, CNNs excel at capturing local time-frequency spectral features, and DNNs provide good baselines with simpler architectures. Including recurrence allows us to evaluate the impact of temporal dependencies on robustness under noisy and out-of-distribution conditions.

Our results offer a valuable resource for researchers and practitioners, enabling informed choices of SCR models not solely based on their number of parameters or their accuracy, but also on their empirical robustness and generalization capabilities in realistic, noisy environments. This approach aims to facilitate the development of more reliable SCR systems that can adapt to diverse and challenging conditions.

\begin{table}[t]
    \caption{Training datasets. SA$\dagger$ means SpecAugment~\cite{park2019specaugment}, which applies time and frequency masking along with speed perturbation.}
    \centering
    \setlength{\tabcolsep}{2pt}
    \ninept
    \begin{tabular}{@{}lcccccc@{}}
    \toprule
    Name         & GSC    & GSC-SA    & GSC-env    & GSC-imp & GSC-env+imp    \\ \midrule
    Alteration & None     & SA$\dagger$       & env noises        & imp noises      & env$+$imp noises    \\
    SNR (dB)     & -      & -      & [-5, 25]    & [-5, 25]    & [-5, 25]    \\
     \bottomrule
    \end{tabular}
    \label{tab:Training datasets}
\end{table}

More than quantifying robustness and general improvement of various models, this work presents a comprehensive benchmark aimed at identifying which models demonstrate the highest robustness for the SCR task across various noise conditions and domain shifts. By systematically comparing models with different complexities and architectures, we provide insights that go beyond simple accuracy or parameter counts.


\section{Experimental Setup}

\subsection{Models}

We experiment with five model architectures, namely DNN, DNN-GRU~(a DNN followed by GRU~\cite{GRU} layers), CNN, CNN14~\cite{kong2020panns}, and TDNN~\cite{speechbrain_gsc_recipe}. Each model is trained multiple times using different depths~(number of layers) and widths~(number of neurons), leading to a total of 40 models of various sizes~(between 20k and 37M parameters). We refer to appendix~\ref{appendix:architectures} for more details. 
This design allows us to assess the effect of both depth and width on model performance and robustness, along with the impact of recurrent layers.

All experiments were carried out using the SpeechBrain toolkit~\cite{speechbrain}. Each model was trained for 50 epochs using the Adam~\cite{kingma2017adam} optimizer. We used a linear scheduler to anneal the learning rate, and early-stopping to choose the best checkpoint based on the validation error rate. The other training hyperparameters are summarized in Table~\ref{tab:hyperparameters}. In total, we trained more than 300 models across all our experiments, with a cumulated training time of more than 200 GPU-hours.



\subsection{Training datasets}

All models are trained exclusively on Google Speech Commands (GSC) \cite{warden2018speech}, an open-source corpus of short~(1-second) commands. We experimented on five dataset variants corresponding to different data augmentation strategies that are detailed in Table~\ref{tab:Training datasets}. The corpus~($\approx$100k samples) was split into train, validation, and test sets with 89/01/10 proportions. The resulting train set is composed of 90,658 files, or 25 hours of audio. Validation and test splits each contain unique speakers not seen during training.


To create noisy training data, we mixed all training samples with additive noise. 
The noise samples used for training consist of 10 environmental sounds, each several minutes long, and 9 impulsive sounds. 
They respectively correspond to vehicle/motor sounds and explosions/gunshots. All noises come from a private database of audio recorded in realistic conditions. Noisy train sets were created as follows: for each clean utterance, we sampled a random segment of environmental and impulsive noises of the same length, along with a random signal-to-noise ratio~(SNR) between -5 and 25dB\footnotemark. These segments were mixed with the utterance to form the GSC-env and GSC-imp sets. The same segments and SNR were then reused to form the combined GSC-env+imp set, yielding three parallel noisy training sets with consistent noise alignment and SNR, for controlled comparisons between noise types.

\footnotetext{This range reflects realistic acoustic conditions, from highly degraded environments to relatively clean scenarios with moderate background noise. This allows us to train models to handle both extreme and mild noise levels.}


\subsection{Testing datasets}


\begin{table}
    \caption{Testing datasets. ID Speech$\dagger$ means that the utterances come from the same domain that was seen during training. Similarly, ID Noise$\ddagger$ means that the noise samples used to corrupt the test set are similar to those of noisy train sets. NS stands for Noise Seen, NU for Noise Unseen, and NM for Noise MUSAN~\cite{snyder2015musan}.}
    \centering
    \setlength{\tabcolsep}{4pt}
    \begin{tabular}{@{}lcccccc@{}}
    \toprule
    Name         & GSC    & GSC-NS    & GSC-NU    & GSC-NM & CV     & TI    \\ \midrule
    \# commands & 35     & 35        & 35        & 35      & 12     & 10    \\
    ID Speech$\dagger$    & ✓      & ✓         & ✓         & ✓       & ✗      & ✗     \\
    ID Noise$\ddagger$     & -      & ✓         & ✓         & ✗       & -      & -     \\
    Noise        & -      & env$+$imp & env$+$imp & MUSAN   & -      & -     \\
    SNR (dB)     & -      &5      & 5     & 5    & -      & -     \\
    \# audios    & 11k & 11k    & 11k    & 11k  & 28k & 4k \\ \bottomrule
    \end{tabular}
    \label{tab:Testing datasets}
\end{table}

The models are evaluated on the test set of GSC along with five other sets representing various distribution shifts. We summarize their characteristics in Table~\ref{tab:Testing datasets}. All the sets derived from GSC are based on the same audio files. For the Common Voice~(CV)~\cite{ardila2020commonvoice} and Texas Instruments~(TI)~\cite{liberman1993ti46} sets, we filtered the commands that matched those of GSC: the digits, "yes", and "no" for CV, and only the digits for TI. For the noisy test sets, we chose a unique SNR of 5dB for all samples because it provides a realistic setting without excessively increasing the experimental load.

\subsection{Features}

We also investigated how the choice of input features affects the model's robustness to OOD data. To do so, we trained our five original architectures using four types of features, which were used as input to the models described above: raw waveforms without any transformation, 64-coefficient mel spectrograms, 20-dimensional MFCCs, and HuBERT~\cite{HuBERT} SSL features. We used a base pre-trained model along with a large model fine-tuned for ASR. Each pre-trained model was frozen, and we experimented with both the last layer and one intermediate Transformer layer that may encode more linguistic information than the final layer~(see prior work~\cite{vielzeuf2024investigating}).

\subsection{Experiments evaluation}
\label{subsec:metrics}

Rather than simply reporting OOD accuracy and comparing it across different experiments, we found it more insightful to introduce two complementary metrics that quantify the potential improvement of a given training condition over the baseline. These metrics allow us to distinguish between overall performance gains and improvements in robustness. We define them as Fairness (F) and Robustness (R).

Fairness (\(F\)) measures the overall accuracy improvement between two training conditions. We used it to compare the results from a given setting with the results from the clean training. It is computed as:
\begin{equation}
    F = \frac{1}{N}\sum_{i=1}^{N}(x'_i - x_i) + (y'_i - y_i),
\end{equation}
where $N=40$ is the number of models, \(x_i\) and \(y_i\) are the models' accuracies on the clean and noisy/OOD test sets for the baseline setting, and idem \(x'_i, y'_i\) are the accuracies for another setting.

Robustness (\(R\)) quantifies how much the models approach the ideal scenario where the ID and OOD accuracies are equal (i.e. \(y = x\)). It is defined as:
\begin{equation}
    R = \frac{1}{N}\sum_{i=1}^{N}\frac{d_i}{d'_i} - 1
\quad\mathrm{and}\quad
    d_i = \frac{|x_i - y_i|}{\sqrt{2}},
\end{equation}
where $d_i=|x_i - y_i|/\sqrt{2}$ is the distance of the baseline models from the ideal $y=x$ line, and \(d'_i\) is the same for the other training condition. 

For each training condition, we calculate the mean of $F$ and $R$ across models to quantify how much the training improved the average model performance compared to the clean setting. Accuracy alone does not reveal whether performance gains come from an overall improvement across all settings or better out-of-distribution generalization, which is why complementary metrics like Fairness and Robustness are needed to disentangle these effects. In addition, these metrics are also useful to summarize the scatter plots of our results, informing us on the general performance trends across the models. Indeed, analyzing 40 ID and OOD accuracies per condition is both overwhelming and noisy, given the stochastic nature of training.




\section{Results}

\begin{figure}[!t]
    \centering
    \includegraphics[width=1\linewidth]{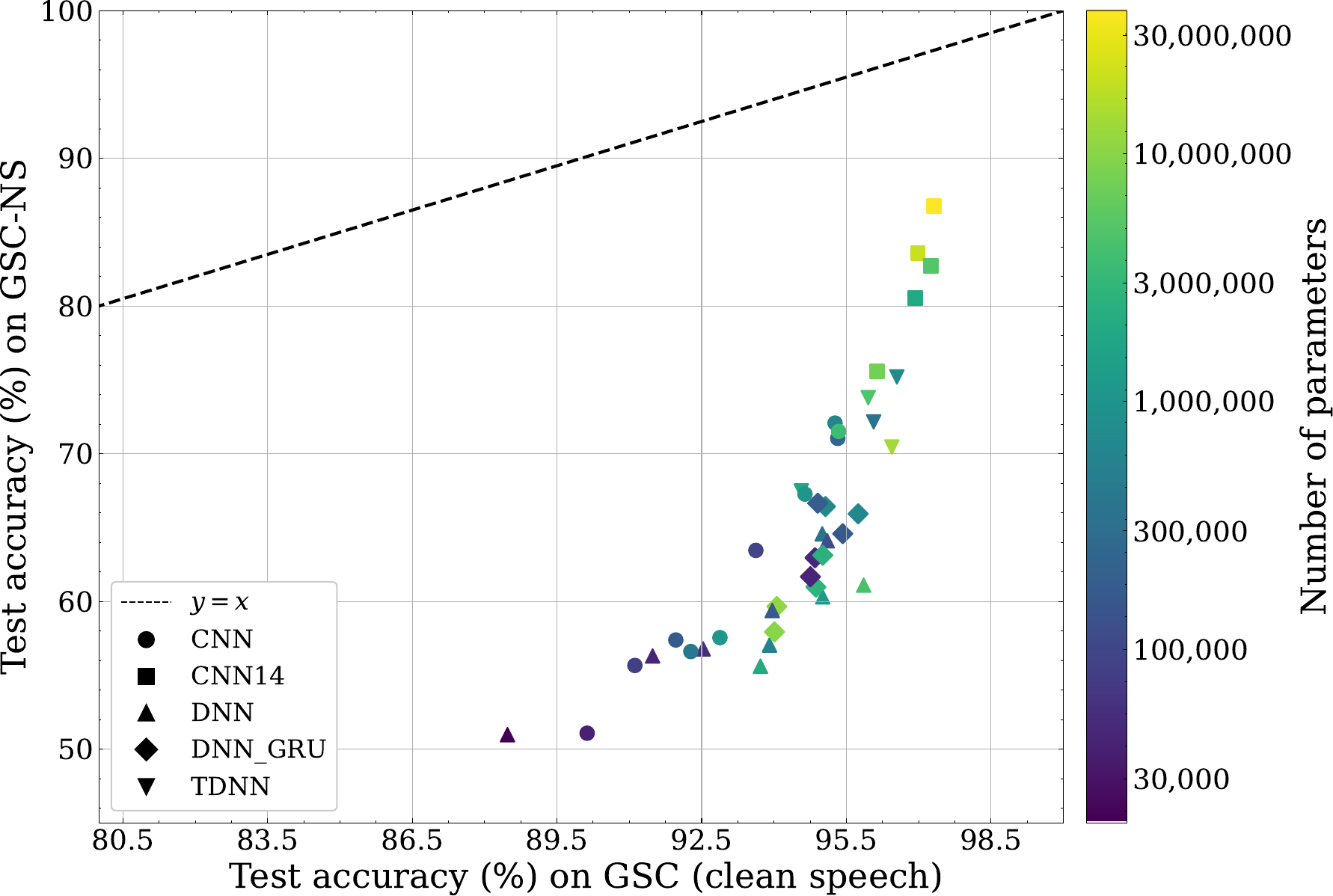}
    \caption{Correlation between ID and OOD accuracies across architectures and sizes. Models were trained on mel spectrograms extracted from clean data.}
    \label{fig: Mel clean and nb param}
\end{figure}

We begin by examining the ID-OOD accuracy correlation for models trained on clean speech and evaluated on both clean and noisy speech. We observe in Figure~\ref{fig: Mel clean and nb param} that there is a nice correlation between ID and OOD performance, especially for the thirty best-performing models. At lower accuracies, the linear relation begins to break down, recalling the "\emph{accuracy on the curve}" phenomenon~\cite{accuracy_curve}.

Additionally, we can see that the model size is not a good predictor of performance; though there seems to be a logarithmic trend with larger models having better accuracies, smaller or simpler models may also outperform larger ones even under noisy conditions.


\subsection{Impact of noisy training data}

\begin{figure}[!t]
    \centering
    \includegraphics[width=1\linewidth]{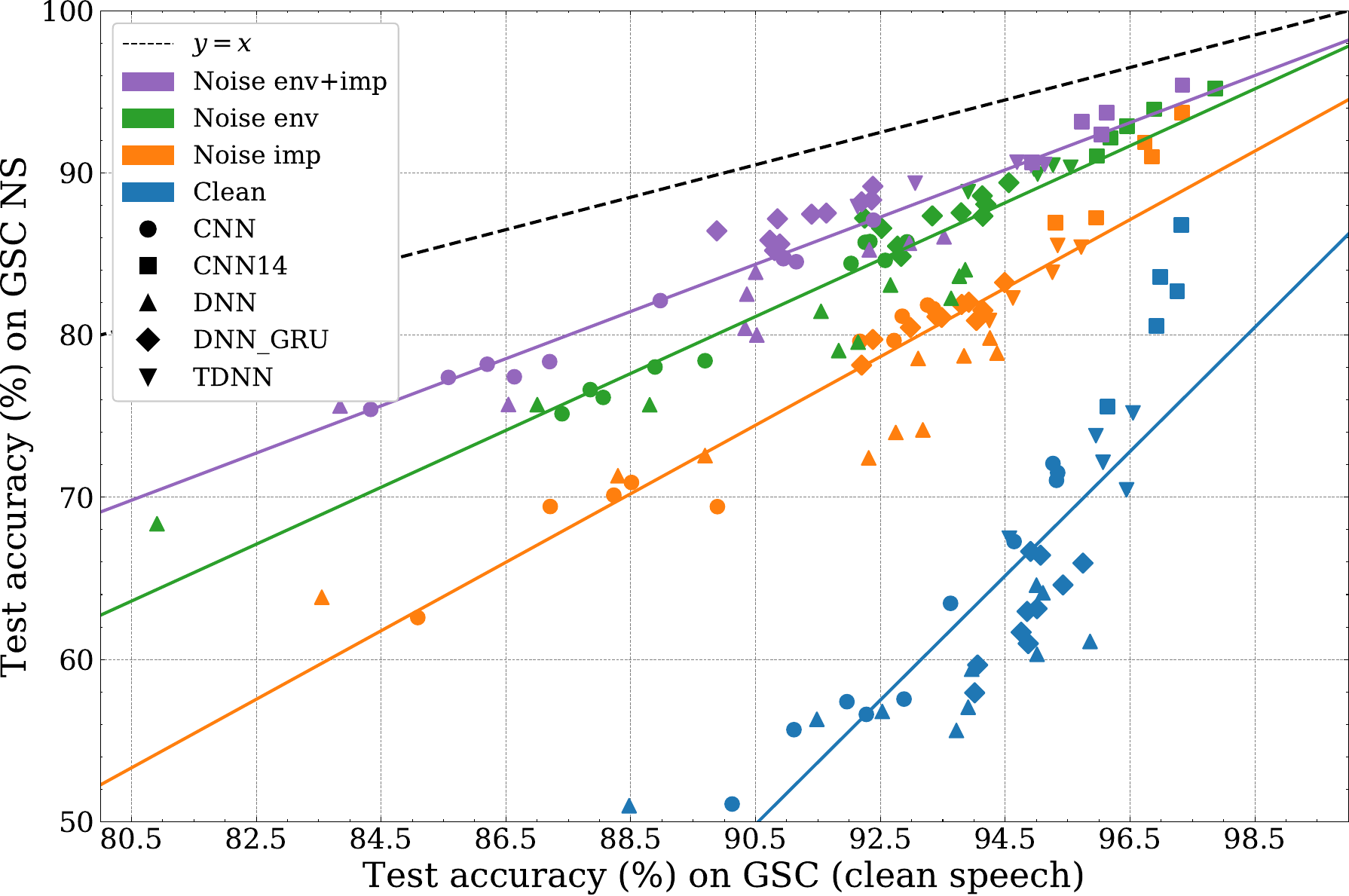}
    \caption{Effect of noise-aware training on clean and noisy accuracies. Models are trained using mel spectrogram features.}
    \label{fig: comparaison training}
\end{figure}

\begin{figure}
    \centering
    \includegraphics[width=1\linewidth]{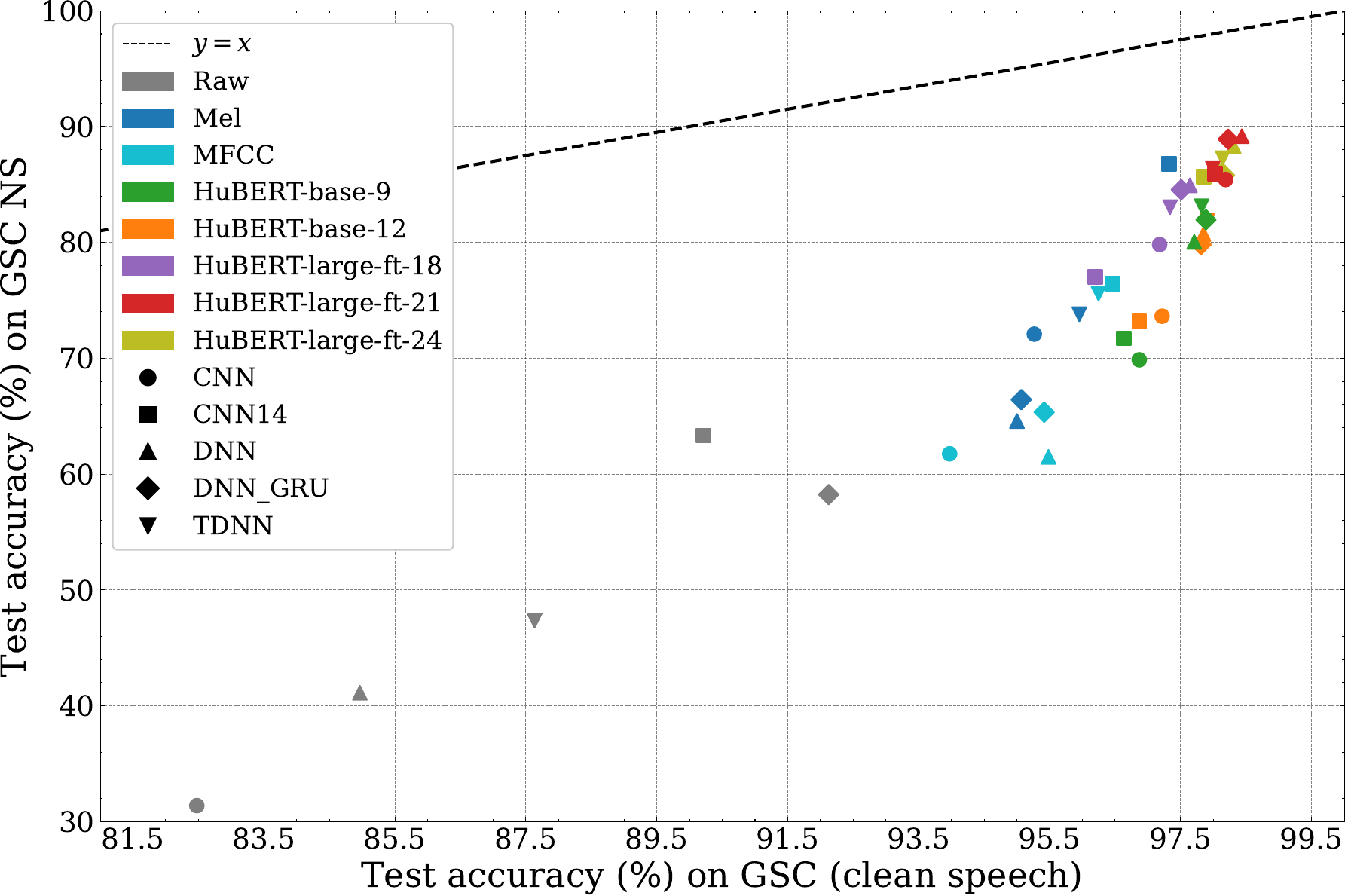}
    \caption{Effect of input features on the clean and noisy accuracies. Models are trained on clean data.}
    \label{fig: comparaison features}
\end{figure}

Next, we train our models on the noisy datasets~(Table~\ref{tab:Training datasets}) and evaluate them on the clean, noisy, and OOD test sets~(Table~\ref{tab:Testing datasets}). Results for GSC-NS are shown on Figure~\ref{fig: comparaison training}; results for other conditions are summarized in Table~\ref{tab: R and F value}.

As expected, we observe that training on noisy data~("noise-aware training") leads to improved performance on noisy test sets. Environmental noise training yields an $F$-score of 0.18 on the GSC-NS set and 0.05 on MUSAN, while impulsive noise achieves lower performance (GSC-NS: 0.12, MUSAN: -0.008), showing that environmental noise contributes more to generalization. We can also see that the combination of both noise types achieves the highest $R$ scores across the settings, reaching 4.38 and 2.95 on the ID-noise tests. This means that on average, models are approximately three times closer to the $y=x$ line when trained on noise, suggesting a much better alignment between clean and noisy performance. In addition, after running linear regressions, we observe that the ID-OOD correlations are much stronger~(see Figure~\ref{fig: comparaison training}) when training on noisy ($R^2 > 0.87$) rather than clean ($R^2 = 0.71$) speech. The slopes of the regression lines are also much smaller, meaning that a small drop in clean speech accuracy does not lead to a dramatic decrease on noisy data.  



When evaluated on TI and CV test sets, noise-aware models underperform the baseline~(negative $F$-scores). $R$-scores on TI are also negative, meaning that noise-aware models are less robust on that set; however, these results should be taken with a grain of salt, considering that the accuracy on TI is very high to begin with~($>90\%$ for the clean setting).
The results on CV are more interesting: noise-aware models obtain positive $R$-scores, meaning that despite slightly reduced accuracies, models are more robust overall, lessening the gap between seen and unseen speakers.


\subsection{Impact of training features}


Next, we train our models using GSC (clean speech) but with different input features. Results are shown on Figure~\ref{fig: comparaison features}; we only show 5 models per feature type for readability. We find that mel-spectrograms far outperform raw waveforms. Converting mel spectrograms to MFCCs leads to inferior results, which suggests that spectrograms represent a good middle-ground among classical features in terms of complexity.

Switching to pretrained SSL features leads to interesting insights. First, we observe that on average, HuBERT features improve both performance and robustness. However, not all architectures actually benefit from using these complex features; for instance, CNN14 works better with spectrograms than HuBERT-base. Also, we can see that while capacity is important (large models outperform base ones), the precise layer used to extract features does not matter so much, with all layers performing roughly equally.



As a last experiment, we used the best training settings (HuBERT-large-21 and env+imp training noise) to retrain all our models. As we can see on Table~\ref{tab: Hubert with noise}, this leads to considerable improvements in $F$ and $R$ scores on all test sets, meaning that on average, models tend to be much more robust when trained using this configuration. We also report the error rates obtained by the best model for each setting on the first three rows of Table~\ref{tab: Hubert with noise}. While the best results on noisy speech are obtained by combining mel spectrograms with noisy training, training on noise degrades performance on unseen speakers; this can be alleviated by switching to HuBERT features.



\begin{table}
    \caption{Results for noisy training. Metrics $F$ and $R$ are defined in section~\ref{subsec:metrics}. Models are tested on $\dagger$noisy or $\ddagger$OOD test sets.}
    \centering
    \label{tab: R and F value}
    \setlength{\tabcolsep}{4pt}
    \begin{tabular}{@{}lccccc@{}}
        \toprule
        Setting      & GSC-NS$\dagger$   & GSC-NU$\dagger$  & GSC-NM$\dagger$$\ddagger$  & TI$\ddagger$    & CV$\ddagger$    \\ \midrule
        \multicolumn{6}{c}{\textbf{Fairness $F$}} \\ \midrule
        SpecAug.   & 0.05 & 0.05 & 0.03   & \textbf{0.01}  & \textbf{0.006} \\
        Env. noise    & \textbf{0.18} & \textbf{0.18} & \textbf{0.05}   & -0.03 & -0.06 \\
        Imp. noise    & 0.12 & 0.11 & -0.008 & -0.02 & -0.05 \\
        Env. $+$ imp. & 0.16 & 0.15 & 0.01   & -0.05 & -0.09 \\ \midrule
        \multicolumn{6}{c}{\textbf{Robustness $R$}} \\ \midrule
        SpecAug.     & 0.14     & 0.13      & 0.08        & \textbf{0.43}  & -0.07 \\ 
        Env. noise       & 3.05     & 2.44      & \textbf{0.77}        & -0.12 & \textbf{30.41} \\ 
        Imp. noise       & 1.23     & 0.91      & 0.15        & -0.16 & 0.90  \\ 
        Env. $+$ imp. & \textbf{4.38}     & \textbf{2.95}      & 0.65        & -0.27 & 2.20  \\ \bottomrule
    \end{tabular}
\end{table}

\begin{table}
    \caption{First 3 rows: best results (\% Error Rate). Last 2 rows: $F$ and $R$ values for noise-aware training with HuBERT features.}
    \centering
    \label{tab: Hubert with noise}
    \setlength{\tabcolsep}{4pt}
    \begin{tabular}{@{}lcccccc@{}}
        \toprule
              & GSC & GSC-NS   & GSC-NU  & GSC-NM  & TI    & CV    \\ \midrule
        Mel+clean & 2.7 & 13.2 & 15.0 & 12.5 & 0.4 & 2.8 \\
        Mel+noise & 2.1 & \textbf{4.8} & \textbf{5.5} & \textbf{6.1} & 0.5 & 5.4 \\
        HuB.+noise & \textbf{1.5} & 5.3 & 6.3 & 6.8 & \textbf{0.02} & \textbf{2.0}  \\ \midrule
        $F$-value   & - & 0.32 & 0.33 & 0.24   & 0.05  & 0.08 \\ 
        $R$-value   & - & 5.72     & 4.67      & 2.83        & 2.60  & 7.97\\ 
         \bottomrule
    \end{tabular}
\end{table}

\subsection{Discussion}



We find that training on combined noises is better than training on either kind of noise (env or imp) in isolation, which is in turn better than applying SpecAugment, which did not lead to good results except on unseen speakers. We also remark that models perform best on GSC-NS, followed by GSC-NU, then GSC-NM, independently of the training setting or evaluation metric. This confirms a known caveat regarding noise robustness: models should never be evaluated on the same set of noises seen during training, as this would overestimate performance. Simply using different audio files from the same domain leads to more realistic results, and it is clear that using OOD noises (e.g., MUSAN) proves even more challenging, although we are still able to reach good performances on GSC-NM using our best settings.

While we can clearly see the "accuracy-on-the-line" phenomenon in noisy conditions~(Figure~\ref{fig: comparaison training}), the effect is less pronounced for OOD speech~(see Figure~\ref{fig: Common voice, training comparison}).
This may be due to the limited coverage of commands by GSC and TI/CV. We also note that the phenomenon almost completely disappears when using HuBERT features, since in this setting all models reach similar high performances.



\section{conclusion}

We have shown that the "accuracy-on-the-line" phenomenon also appears for the command-recognition task, generalizing findings previously reported in the computer vision domain. Our tests include different types of OOD conditions, such as noise or unseen speakers. We investigate how robustness can be improved by using either data augmentation or more powerful features. One obvious drawback of using HuBERT features is the huge computational cost; we leave the exploration of more efficient neural feature extractors as future work. Interesting follow-ups also include experimenting with other audio tasks such as ASR or speaker recognition, or evaluating the potential of synthetic data to improve speaker robustness.



\vfill\pagebreak


\section{Acknowledgments}

We would like to thank François Capman for his guidance and his help during the writing of this paper.

\bibliographystyle{IEEEbib}
\bibliography{biblio}

\begin{thebibliography}{10}

\bibitem{pmlr-v139-miller21b}
John~P Miller, Rohan Taori, Aditi Raghunathan, et~al.,
\newblock ``Accuracy on the line: on the strong correlation between out-of-distribution and in-distribution generalization,''
\newblock in {\em Proceedings of the 38th International Conference on Machine Learning}, Marina Meila and Tong Zhang, Eds. 18--24 Jul 2021, vol. 139 of {\em Proceedings of Machine Learning Research}, pp. 7721--7735, PMLR.

\bibitem{baek2022agreement}
Christina Baek, Yiding Jiang, Aditi Raghunathan, and J~Zico Kolter,
\newblock ``Agreement-on-the-line: Predicting the performance of neural networks under distribution shift,''
\newblock {\em Advances in Neural Information Processing Systems}, vol. 35, pp. 19274--19289, 2022.

\bibitem{Porjazovski_2024}
Dejan Porjazovski, Anssi Moisio, and Mikko Kurimo,
\newblock ``Out-of-distribution generalisation in spoken language understanding,''
\newblock in {\em Interspeech 2024}. Sept. 2024, pp. 807--811, ISCA.

\bibitem{DBLP:journals/corr/abs-2403-07937}
Muhammad~A. Shah, David~Solans Noguero, Mikko~A. Heikkila, et~al.,
\newblock ``Speech robust bench: A robustness benchmark for speech recognition,'' 2024.

\bibitem{rakib2023ood}
Fazle~Rabbi Rakib, Souhardya~Saha Dip, Samiul Alam, et~al.,
\newblock ``{OOD}-speech: A large {B}engali speech recognition dataset for out-of-distribution benchmarking,'' 2023.

\bibitem{sanyal2024accuracy}
Amartya Sanyal, Yaxi Hu, Yaodong Yu, et~al.,
\newblock ``Accuracy on the wrong line: On the pitfalls of noisy data for {OOD} generalisation,''
\newblock in {\em ICML 2024 Next Generation of AI Safety Workshop}, 2024.

\bibitem{snyder2019speaker}
David Snyder, Daniel Garcia-Romero, Gregory Sell, et~al.,
\newblock ``Speaker recognition for multi-speaker conversations using {X}-vectors,''
\newblock in {\em ICASSP 2019 - 2019 IEEE International Conference on Acoustics, Speech and Signal Processing (ICASSP)}, 2019, pp. 5796--5800.

\bibitem{park2019specaugment}
Daniel~S. Park, William Chan, Yu~Zhang, et~al.,
\newblock ``Spec{A}ugment: A simple data augmentation method for automatic speech recognition,''
\newblock in {\em Interspeech 2019}, 2019, pp. 2613--2617.

\bibitem{GRU}
Kyunghyun Cho, Bart van Merri{\"e}nboer, Dzmitry Bahdanau, and Yoshua Bengio,
\newblock ``On the properties of neural machine translation: Encoder{--}decoder approaches,''
\newblock in {\em Proceedings of {SSST}-8, Eighth Workshop on Syntax, Semantics and Structure in Statistical Translation}, Dekai Wu et~al., Eds., Doha, Qatar, Oct. 2014, pp. 103--111, Association for Computational Linguistics.

\bibitem{kong2020panns}
Qiuqiang Kong, Yin Cao, Turab Iqbal, et~al.,
\newblock ``Panns: Large-scale pretrained audio neural networks for audio pattern recognition,''
\newblock {\em IEEE/ACM Transactions on Audio, Speech, and Language Processing}, vol. 28, pp. 2880--2894, 2020.

\bibitem{speechbrain_gsc_recipe}
{SpeechBrain team},
\newblock ``Google {S}peech {C}ommands recipe,'' \url{https://github.com/speechbrain/speechbrain/tree/develop/recipes/Google-speech-commands}, 2021,
\newblock Accessed: 2025-06-25.

\bibitem{speechbrain}
Mirco Ravanelli, Titouan Parcollet, Adel Moumen, et~al.,
\newblock ``Open-source conversational {AI} with {S}peech{B}rain 1.0,''
\newblock {\em Journal of Machine Learning Research}, vol. 25, no. 333, 2024.

\bibitem{kingma2017adam}
Jimmy Ba and Diederik~P. Kingma,
\newblock ``Adam: A method for stochastic optimization,''
\newblock in {\em International Conference on Learning Representations (ICLR)}, 2015.

\bibitem{warden2018speech}
Pete Warden,
\newblock ``Speech {C}ommands: A dataset for limited-vocabulary speech recognition,'' 2018.

\bibitem{snyder2015musan}
David Snyder, Guoguo Chen, and Daniel Povey,
\newblock ``{MUSAN}: A {M}usic, {S}peech, and {N}oise corpus,'' 2015.

\bibitem{ardila2020commonvoice}
Rosana Ardila, Megan Branson, Kelly Davis, et~al.,
\newblock ``{C}ommon {V}oice: A massively-multilingual speech corpus,''
\newblock in {\em Proceedings of the Twelfth Language Resources and Evaluation Conference}, Nicoletta Calzolari et~al., Eds., Marseille, France, May 2020, pp. 4218--4222, European Language Resources Association.

\bibitem{liberman1993ti46}
Liberman Mark, Amsler Robert, Church Ken, et~al.,
\newblock ``{TI} 46-word [speech corpus] ({LDC93S9}),'' Linguistic Data Consortium, 1993.

\bibitem{HuBERT}
Wei-Ning Hsu, Benjamin Bolte, Yao-Hung~Hubert Tsai, et~al.,
\newblock ``{HuBERT}: Self-supervised speech representation learning by masked prediction of hidden units,''
\newblock {\em IEEE/ACM Transactions on Audio, Speech, and Language Processing}, vol. 29, pp. 3451--3460, 2021.

\bibitem{vielzeuf2024investigating}
Valentin Vielzeuf,
\newblock ``Investigating the 'autoencoder behavior' in speech self-supervised models: a focus on {HuBERT}'s pretraining,'' 2024.

\bibitem{accuracy_curve}
Weixin Liang, Yining Mao, Yongchan Kwon, et~al.,
\newblock ``Accuracy on the curve: On the nonlinear correlation of {ML} performance between data subpopulations,''
\newblock in {\em Proceedings of the 40th International Conference on Machine Learning}, Andreas Krause et~al., Eds. 23--29 Jul 2023, vol. 202 of {\em Proceedings of Machine Learning Research}, pp. 20706--20724, PMLR.

\end{thebibliography}

\vfill\pagebreak
\section{Appendices}
\appendix

\section{Details on model architectures}
\label{appendix:architectures}

A total of 40 models are evaluated, each corresponding to an instance of one of the 5 following architectures: 
\begin{itemize}
\item TDNN: Five layers of dilated 1D convolutions, followed by two fully connected~(FC) layers. Implementation follows~\cite{speechbrain_gsc_recipe}.

\item DNN-GRU: Two FC layers followed by two bi-GRU~\cite{GRU} layers. The last two hidden states of the recurrent network are concatenated and fed into two additional FC layers. 

\item DNN: Three FC layers with context windows of size 3, 3, 1, then a temporal pooling followed by two FC layers.

\item CNN14~\cite{kong2020panns}: Six blocks composed of two 2D convolution layers and one max pooling, then a temporal pooling followed by two FC layers.

\item CNN: Three layers of 2D convolutions with max pooling, then a temporal pooling followed by two FC layers.
\end{itemize}

For CNN, DNN, and DNN-GRU architectures, we evaluated 10 variants of each model.
The first 5 variants retain the original number of layers, but the layer widths are scaled by a factor of ×0.25, ×0.5, ×1 (default), ×2, and ×4.
The remaining 5 variants follow the same scaling strategy but with one layer removed: a convolution layer for CNNs, a fully connected layer for DNNs, and a GRU layer for DNN-GRUs.
This design allows us to assess the impact of both depth and width on model performance and robustness. 

For the TDNN, we evaluate five variants: the default model and four others where the layer width is scaled by a factor of ×0.25, ×0.5, ×2, and ×4.
Finally, for CNN14, we also evaluate five variants: the baseline model, two models with the layer width scaled by ×0.5 and ×0.25, and two additional models using the same scaling factors but with only one repetition of each convolution block (instead of the usual block duplication). 

\begin{table}
    \caption{Training hyperparameters.}
    \centering
    \setlength{\tabcolsep}{4pt}
    \begin{tabular}{@{}llll@{}}
        \toprule
        Sample rate     & 16kHz   & Optimizer       & Adam~\cite{kingma2017adam}                  \\
        FFT size      & 400     & Scheduler       & Linear ($10^{-3}$-$10^{-4}$) \\
        Window size & 400     & Batch size      & 32                    \\
        Hop length      & 160     & Number of epochs & 50                    \\
        \bottomrule
    \end{tabular}
    \label{tab:hyperparameters}
\end{table}

\section{Additional Figures}
\label{appendix:figures}

\begin{figure}
    \centering
    \includegraphics[width=1\linewidth]{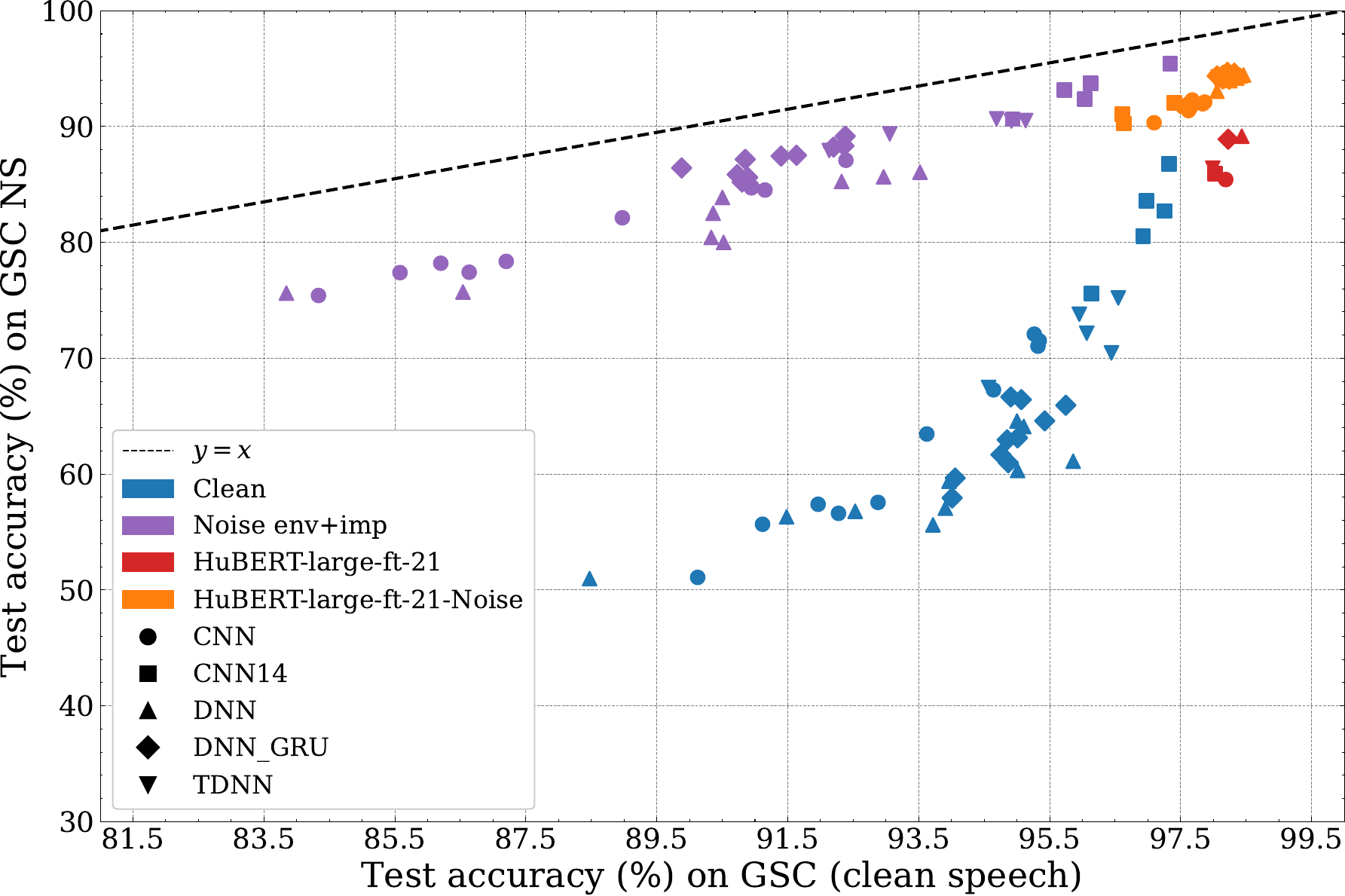}
    \caption{Effect of training with HuBERT features on noisy data compared to mel features.}
    \label{fig: Hubert 21 with noisy training}
\end{figure}

\begin{figure}
    \centering
    \includegraphics[width=1\linewidth]{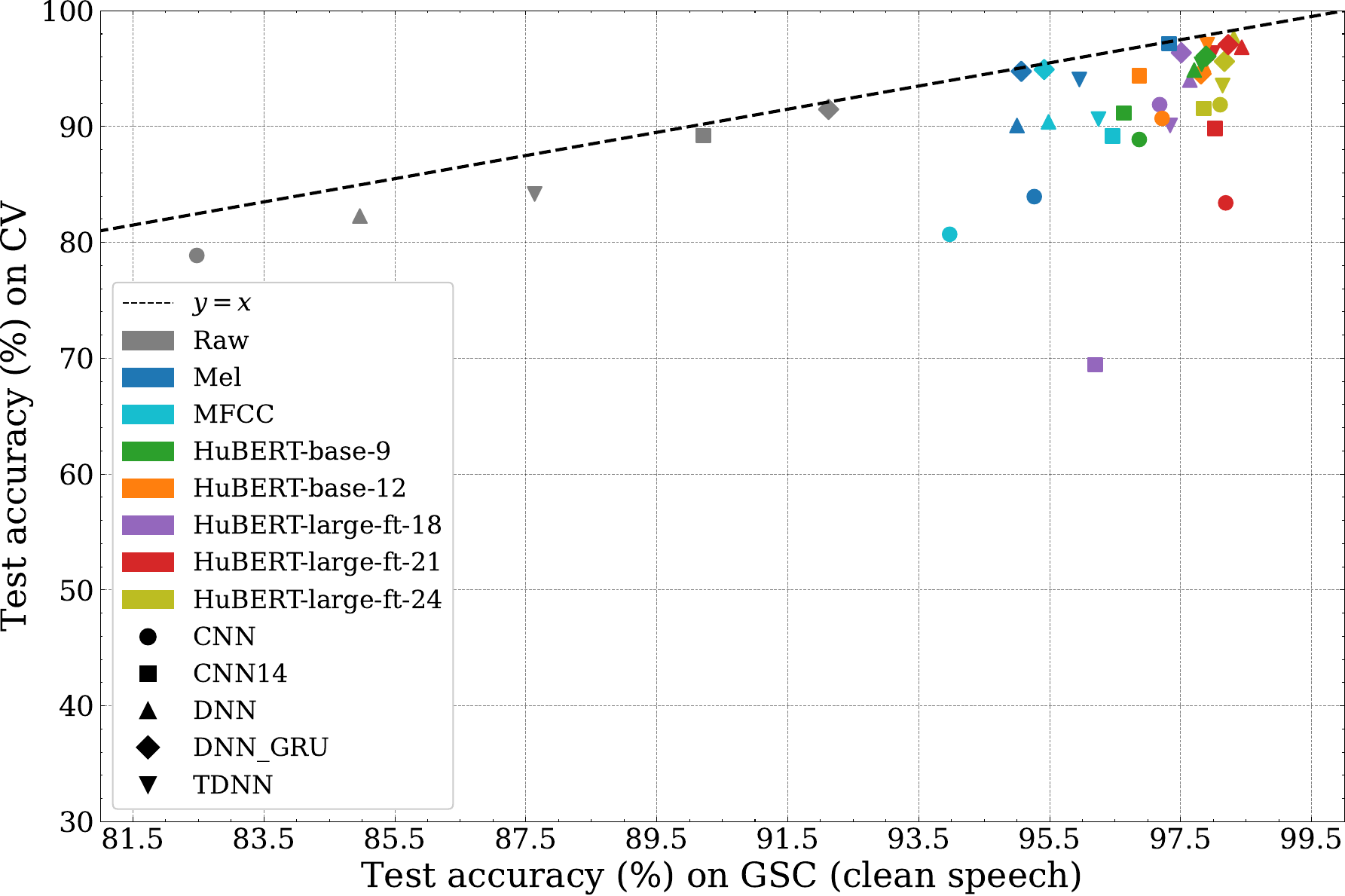}
    \caption{Effect of input features on accuracy for clean and unseen speakers (CV). Models are trained on clean data.}
    \label{fig: Common voice, features comparison}
\end{figure}

\begin{figure}
    \centering
    \includegraphics[width=1\linewidth]{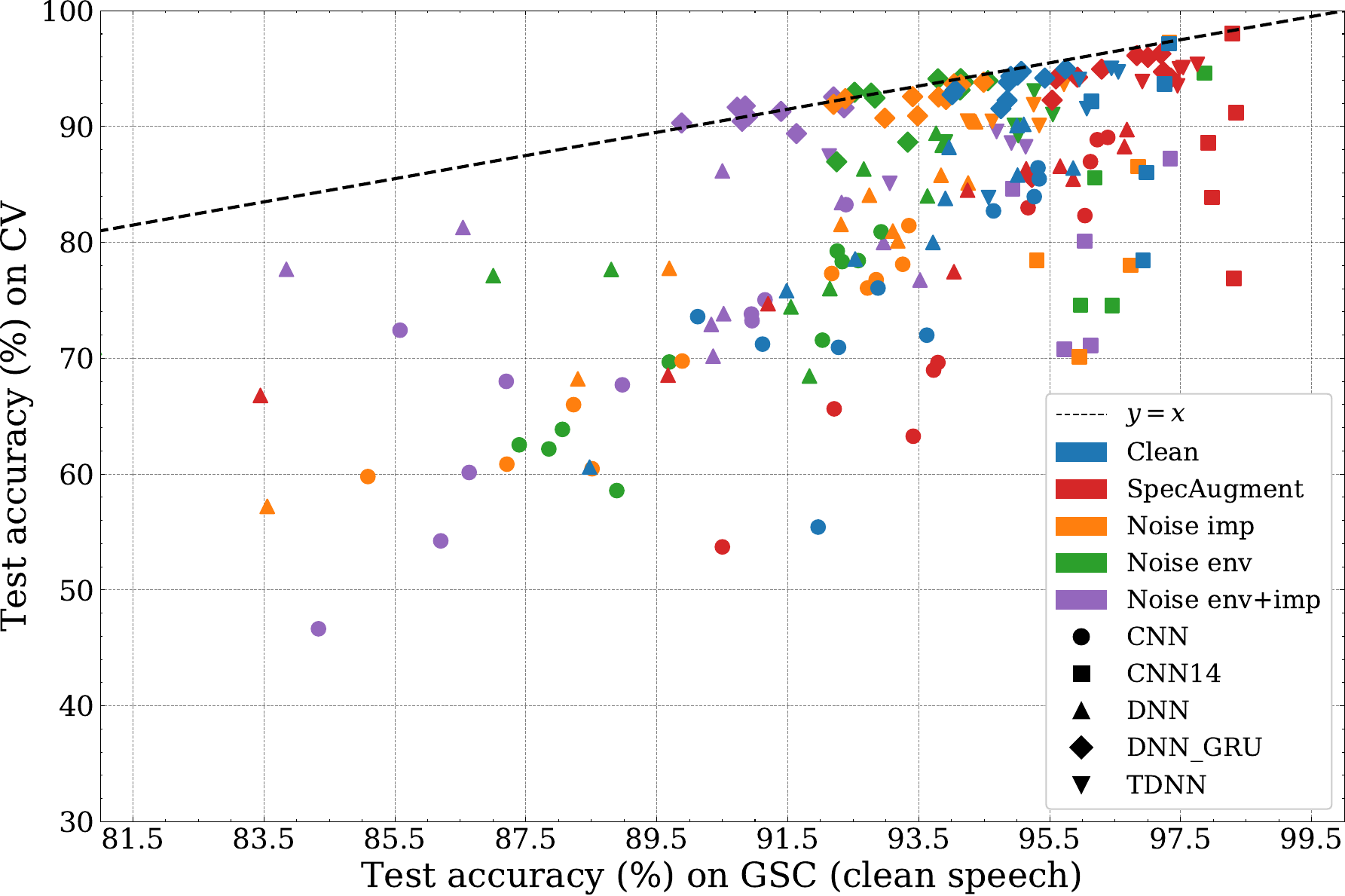}
    \caption{Effect of training conditions on accuracy for clean and unseen speakers (CV). Models use mel features as input.}
    \label{fig: Common voice, training comparison}
\end{figure}

\end{document}